\documentclass{article} 
\usepackage{nips15submit_e,times}
\usepackage{hyperref}
\usepackage{url}
\usepackage{amsmath}
\usepackage{placeins}
\usepackage{graphicx}
\usepackage{float}
\usepackage[usestackEOL]{stackengine}
\setstackgap{L}{\normalbaselineskip}

\title{NFL Play Prediction}
\author{
Brendan Teich* \hspace{1cm} Roman Lutz* \hspace{1cm} Valentin Kassarnig*\\
College of Information and Computer Sciences\\
University of Massachusetts Amherst\\
\texttt{\{bteich, romanlutz, vkassarnig\}@cs.umass.edu} 
}
\nipsfinalcopy % Uncomment for camera-ready version

\usepackage{tabularx}
\usepackage{hyperref}

\begin{document}
\maketitle
\renewcommand{\thefootnote}{\arabic{footnote}}
\let\thefootnote\relax\footnotetext{* The authors contributed equally to this work.}
\newcommand{\thefootnote}{\arabic{footnote}}

\begin{abstract}
Based on NFL game data we try to predict the outcome of a play in multiple different ways including Decision and Classification Trees, Nearest Neighbors, Naive Bayes, Linear Discriminant Analysis, Support Vector Machines and Regression, and Artificial Neural Networks. An application of this is the following: by plugging in various play options one could determine the best play for a given situation in real time. While the outcome of a play can be described in many ways we had the most promising results with a newly defined measure that we call \textit{progress}. We see this work as a first step to include predictive analysis into NFL playcalling.
\end{abstract}

\section{Introduction}
Game strategy and playcalling are an integral part of American Football. Especially at the professional level coaches spend most of their time either analyzing the opponent's past games to find out about weaknesses or teaching their players how to take advantage of the opponent's weaknesses. Even though American Football is a very data-driven sports coaches have the final say over playcalling. This often leads to controversy over whether coaches made good or bad decisions. A prime example is this year's Superbowl where the Seahawks went with a pass instead of a run and ended up getting intercepted and loosing the game \footnote{http://bleacherreport.com/articles/2350553-questionable-play-call-costs-seahawks-super-bowl-victory}. We argue that computers could assist coaches or even completely take over playcalling duties by predicting the outcome of specific plays.\\
For such an application we build the foundation by providing predictions for the outcome of a play based on the game situation and the play description. As a next step coaches could vary the play description using plays from their own playbook and see which play has the best chance of succeeding.\\
This report is structured as follows: Section~\ref{sec:related} looks at related work on play prediction. In Section~\ref{sec:dataset} we describe the data set we used and how we obtained it. We examine the features of the data set in greater detail in Section~\ref{sec:features}. In Section~\ref{sec:targets} we identify preferable targets for our predictions, namely a binary success variable, the number of yards gained, and a newly defined real-valued progress measure. Furthermore, we describe the various methods we used for classification and regression on the outcome of plays in Section~\ref{sec:methods} before evaluating and comparing the results in Section~\ref{sec:eval}. Once the foundation is established, we give an example of how this work could be used in Section~\ref{sec:applications}. Finally, Section~\ref{sec:conclusion} concludes with a summary and future outlook.

\section{Related work}
\label{sec:related}
Stilling and Critchfield~[6] used generalized matching equations to analyze the relationship between play selection and yards gained. They found that undermatching, which may result from the tendency of coaches to "mix up" plays, a bias towards rushing plays, and that the generalized matching equation accounted for a majority of the variability in play selection. Using specialized equations for each down revealed that first down is biased towards rushing, and later downs are biased towards passing. It was also found that rushing was preferred when fewer than 4 yards remained to first down, and that passing was preferred when more than 10 yards remained. Kovash and Levitt~[7] analyzed whether in American Football and Baseball decisions conform with Minimax. By studying 125,000 NFL plays from 2001-2005 they found negative serial correlation and thus predictable tendencies of coaches in calling run or pass plays. Specifically, their results indicate that playcalling should focus more on passing - a conclusion that is inconsistent with Minimax Theory and that at the same time is the basis for the main application of this work. Mitchell~[8] enhanced the matching pennies model by Kovash and Levitt~[7] by adding the notion of investment to playcalling. This helps explain the commonly accepted ideas of "running to wear down a defense" and "running to set up the pass". Reed, Critchfield and Martens~[9] analyzed playcalling with the Generalized Matching Law, a mathematical model of operant choice. They discovered a bias for calling rushing plays and undermatching in the sense of imperfect playcalling. McGarrity and Linnen~[5] developed a game theoretic model to analyze how a team changes its play calling when the starting quarterback is replaced by its backup. Replacement quarterbacks are usually less-experienced which means passing the ball is of higher risk. The paper refers to standard optimization theories which suggest that the team should run more often since in contrast to the quarterback the productivity of the running back has not changed. However, their findings say teams will not change their play calling because the defense will expect more run plays and will consequently play more often against them. That is, quarterback substitutions have less impact than expected. That supports our approach of considering only team-based features and disregarding features based on individuals.\\
Molineaux, Aha and Sukthankar~[11] used plan recognition to identify the defensive strategy and thus improve the results of their case-based Q-learning algorithm. They evaluated this on the open-source American Football simulator Rush 2008. While this is an abstraction from real football at the moment they are working on making it more realistic in the future for their purposes.\\
Lutz~[21] already identified related work some of which we refer to in the following.
Most research in sports prediction focuses on predicting the winner of a match instead of predicting individual plays. Min et al.~[15] used Bayesian inference and rule-based reasoning to predict the result of American Football matches. Their work is based on the idea that sports is both highly probabilistic and at the same time based on rules because of team strategies. Sierra, Fosco and Fierro~[14] used classification methods to predict the outcome of an American Football match based on a limited number of game statistics excluding most scoring categories. In their experiments, linear Support Vector Machines had by far the most accurate predictions and ranked above all expert predictions they compared with. Similarly, Harville~[16] used a linear model for the same task. A Neural Network approach to predicting the winner of College Football games was proposed by Pardee~[17]. His accuracy of $76\%$ improved on expert predictions. Purucker~[13] and Kahn~[12] applied neural networks to NFL game predictions and mostly improved on existing accuracies. A problem of both their results is the small test set since they only evaluated on a total of two weeks of NFL games. Stefani~[18] used an improved least squares rating system to predict the winner for nearly 9000 games from College and Pro Football and even other sports.
Fokoue and Foehrenbach~[19] have analyzed important factors for NFL teams with Data Mining. Their results are especially useful for this work because the first step for predicting plays involves identifying important features.

\section{Data set}
\label{sec:dataset}
The API \textit{nflgame}~[2] provides access to data from all NFL games of the last six years (2009 - 2014) on a play-by-play basis. From this data we extracted a total of 12 features and filtered out irrelevant plays which left us with 177245 plays. The following sections describe this process and the features in greater detail.

\section{Features}
\label{sec:features}

\subsection{Feature extraction}
\label{sec:feature-extraction}
For each play we receive a data structure from the API. This structure contains basic information about the current situation of the game like the game clock, the field position, the possessing team, etc. It also contains a string which describes what happened on the field. Apart from the actual play this string can describe injuries, timeouts, quarterback substitutions, penalties, or other information. Since we are only interested in the actual plays we need to filter out all the irrelevant information. That is, we need to find the sentence which describes the actual play. In cases of penalties we reject the entire string because a penalty impacts the course of the play. Furthermore, there are plays which are of no interest for us which will be also ignored. These include field goals, punts, sacks, fumbles, etc. The final string that describes the current play will then be parsed to extract the features. Table~\ref{tab:feature_table} lists and describes all the features we use.

\begin{table}
\begin{tabularx}{\textwidth}{ | l | l | X | }
  \hline
  \textbf{Feature} & \textbf{Type} & \textbf{Description}\\ \hline
  Team & Categorical & The name of the team which has the ball \\ \hline
  Opponent & Categorical & The name of the team which is defending \\ \hline
  Half & Continuous & Describes whether it is the first or the second half of the game \\ \hline
  Time & Continuous & Remaining time (in seconds) in the current half  \\ \hline
  Field position & Continuous & Distance to the opponent's end zone (in yards) \\ \hline
  Down & Continuous & The current down (1, 2, 3 or 4) \\ \hline
  To-go & Continuous & The remaining distance to the next first down (in yards) \\ \hline
  Shotgun & Binary & Flag, whether the team starts the play in the shotgun formation \\ \hline
  Pass & Binary & Flag, whether the quarterback passes the ball \\ \hline
  Side & Categorical & Side to which the quarterback passes or the rusher runs (left, middle, or right) \\ \hline
  Pass length & Categorical & In case of a pass it describes whether it is a short or a deep pass. Otherwise it is 0. \\ \hline
  QB run & Binary & Flag, whether the quarterback runs himself \\ \hline
\end{tabularx}
\caption{List of play features}
\label{tab:feature_table}
\end{table}

\subsubsection{Examples}

Table~\ref{tab:feature_examples} in the Appendix shows some example play descriptions and the extracted features and labels. Omitted features have value 0. Note that not all features are extracted from the description string. Some features are directly obtained from the data structure obtained from the API. The table also lists the different labels which are explained in section~\ref{sec:targets}.

\subsection{Encoding}
\label{sec:encoding}

Some of our features have categorical values. For example either of the features $Team$ and $Opponent$ take on a value representing one of the 32 teams. Simply numbering the teams from 1 to 32 would not represent the real situation because team \#1 is not closer to team \#2 than it is to team \#32. Consequently, we need a more sophisticated encoding. For that reason, we have chosen one-hot encoding for our categorical features. That is, each categorical feature is replaced by $k$ binary features where $k$ is the number of possible values. For example, the feature "team" is replaced by 32 features like $team=GB$, $team=NE$, and so on. In each sample only one of these features has the value "1" while all others are "0".
Encoding all our categorical features of our data set expands the size from 12 to 77 dimensions.

\subsection{Exploratory data analysis}
\label{sec:dim-red}
\begin{figure}[!tbp]
  \centering
  \begin{minipage}[b]{0.47\textwidth}
    \includegraphics[width=\textwidth]{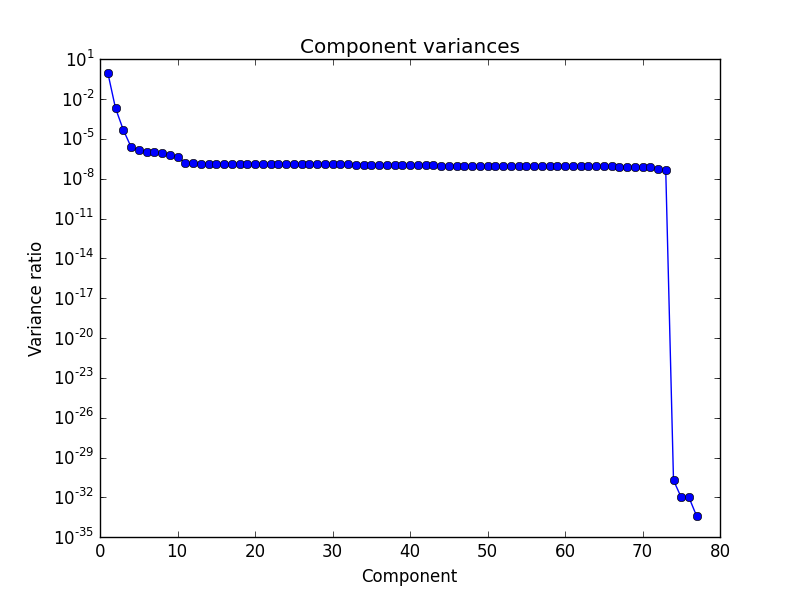}
    \caption{Variance ratios of components}
    \label{fig:variance}
  \end{minipage}
  \hfill
  \begin{minipage}[b]{0.50\textwidth}
    \includegraphics[width=\textwidth]{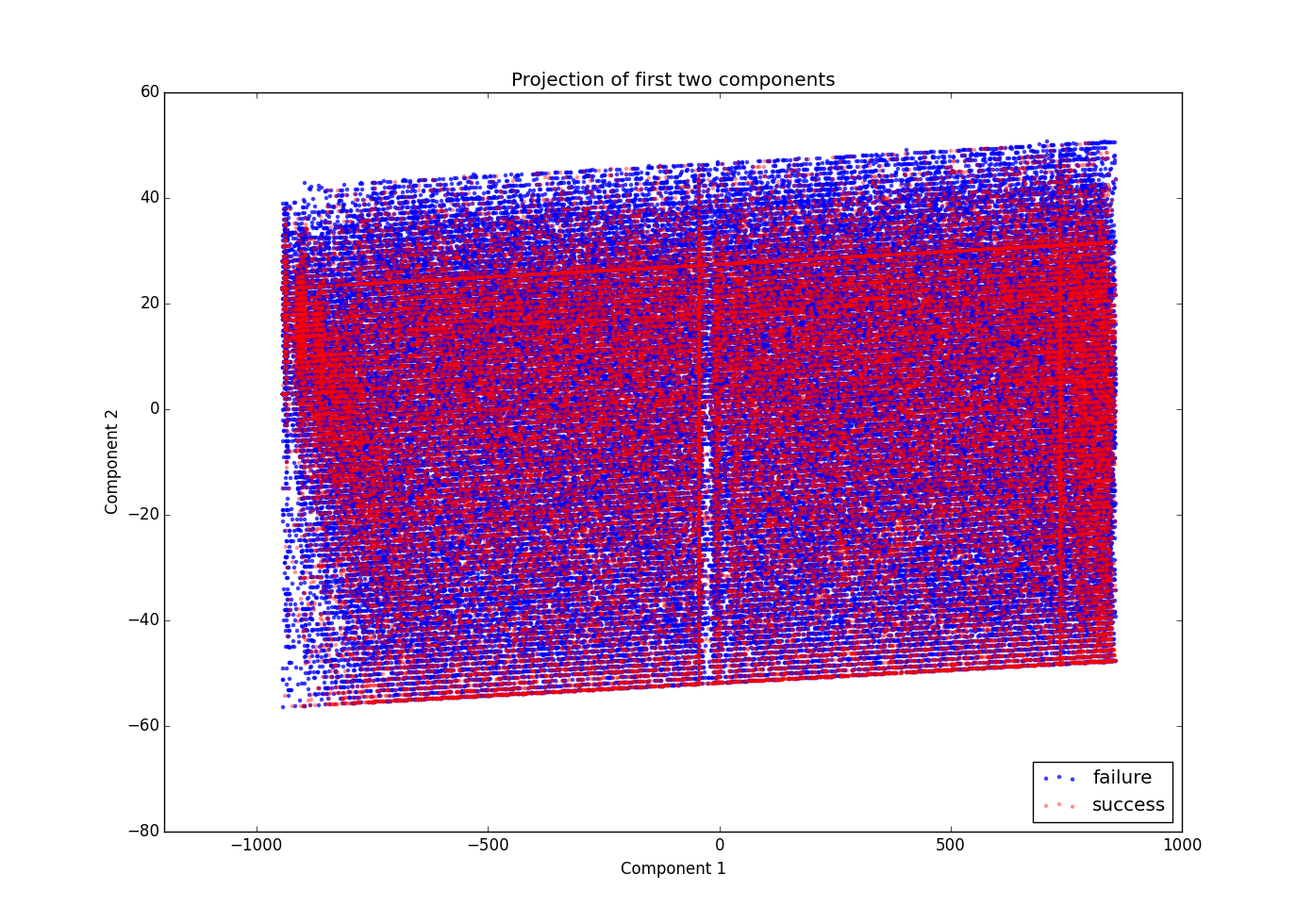}
    \caption{Projection on the first two components}
  \label{fig:2dproj}
  \end{minipage}
\end{figure}

In order to analyze the distribution of data we performed a principal component analysis (PCA). Figure~\ref{fig:variance} shows a plot of the variance ratios of the resulting components. The first component already covers more than 99.7\% of the variance. 
Figure~\ref{fig:2dproj} shows a projection of the entire data set on the two dimnsions that represent most of the variance. Red points represent successful plays and blue points are failure plays. The distribution of the plays shows some sort of pattern or structure. Noticeable are the two red clusters close to the left and the right edge. However, the data is far too noisy to separate the two classes.

Note that PCA relies on continuous-valued variables while our data consists of a mixture of continuous and binary values. As discussed by Yu and Tresp [20] there are some issues when applying PCA to mixed types of data. Hence, in our case PCA helps us to explore the data but it is not suitable for a dimensionality reduction. A different approach of reducing the dimensionality is to just remove features which have only very little impact on the classification. One method is to calculate the analysis of variance (ANOVA) F-test statistic and rank the features according to their score. The ANOVA F-Test statistic is the ratio of the average variability among groups to the average variability within groups. That is, it determines whether the means between the two groups are significantly different.
Table~\ref{tab:feature_scores} in the Appendix shows the computed F-values of each feature. It can be seen that the feature for the remaining distance to the next first-down ($togo$) has by far the highest score. An interesting observation is the wide range of scores of team-specific features. For example the feature for the New England Patriots ($team=NE$) has a score which is around a million times higher than for the Miami Dolphins ($team=MIA$). This gives evidence that some teams might be more predictable than others. \\
%This will be further explored in section~X where we train models for individual teams.

\section{Targets}
\label{sec:targets}
Apart from the features we also have to extract the ground truth labels of each play. Since we pursue multiple approaches we also have to extract multiple labels. Table~\ref{tab:label_table} lists and describes all the labels we use.

\begin{table}[h]
\begin{tabularx}{\textwidth}{ | l | l | X | }
  \hline
  \textbf{Label} & \textbf{Type} & \textbf{Description}\\ \hline
  Success & Binary & Indicates whether a play resulted in a first down or a touchdown.  \\ \hline
  Yards & Continuous & Indicates how many yards a team has gained through this play. \\ \hline
  Progress & Continuous & A newly defined measure for progress based on the current down, achieved yards and yards to go. \\ \hline
\end{tabularx}
\caption{List of play labels}
\label{tab:label_table}
\end{table}

While the prediction of \textit{success} is a classification task, \textit{yards} and \textit{progress} are real numbers requiring regression methods. We examine various methods for classification and regression in Section~\ref{sec:methods}.\\
When we use the binary value \textit{success} we define successful plays as plays which either obtain a first down or score a touchdown. In all other cases the play will be classified as failure. With this target definition our data set is quite imbalanced since it contains about 70\% unsuccessful and only 30\% successful samples. That imbalance needs to be taken into account at the different classification methods or else a classifier that classifies each sample as "failure" would achieve a classification accuracy of 70\% which is actually quite good. For that reason we do not only evaluate the accuracy but also the recall and the precision. Those are two error measures which evaluate the quality of the classification. The recall tells how many of the successful samples were classified correctly, while the precision tells how many of the \textit{success} predictions were actually correct. Reconsidering the case were all samples are predicted as "failure", this would yield a precision and recall of 0\%. So, although the accuracy is quite high the quality of the classification is pretty poor. A good way to combine recall and precision is the $F_1$ score which is the harmonic mean of those two values. The $F_1$ score is calculated as shown in the following equation:
\[F_1=2 \cdot \frac{precision \cdot recall}{precision+recall}\]
Still, we argue that both \textit{success} and \textit{yards} are unable to capture whether the result of a play is actually desirable. For example, a 9 yard gain on 1st down and 10 yards to go is given the same label (0) as a 10 yard loss. \textit{success} can not represent this difference. \textit{yards} has similar issues. Imagine a team has a fourth down and 28 yards to go, but achieves only 27 yards. In most situations 27 yards is a great result for a play, but not here since the play would result in a turnover on downs. \textit{yards} would still get the very good value of 27.\\
As a consequence of these observations we invented a completely new success measure which we call \textit{progress}. The special thing about this label is that also takes the current down into account. It is calculated as follows with $down$ representing the current down, $togo$ being the remaining yards to go for a first down and $gained$ representing the yards gained on the play:
\[
progress(down, togo, gained) = \begin{cases}
               0               & down \in \{3, 4\} \text{ and } gained < togo\\
               \left(\dfrac{gained}{togo}\right)^{down}               & down \in \{1, 2\} \text{ and } gained < togo\\
               1 & gained \geq togo
           \end{cases}
\]
\begin{figure}
\centering
  \includegraphics[width=0.75\textwidth]{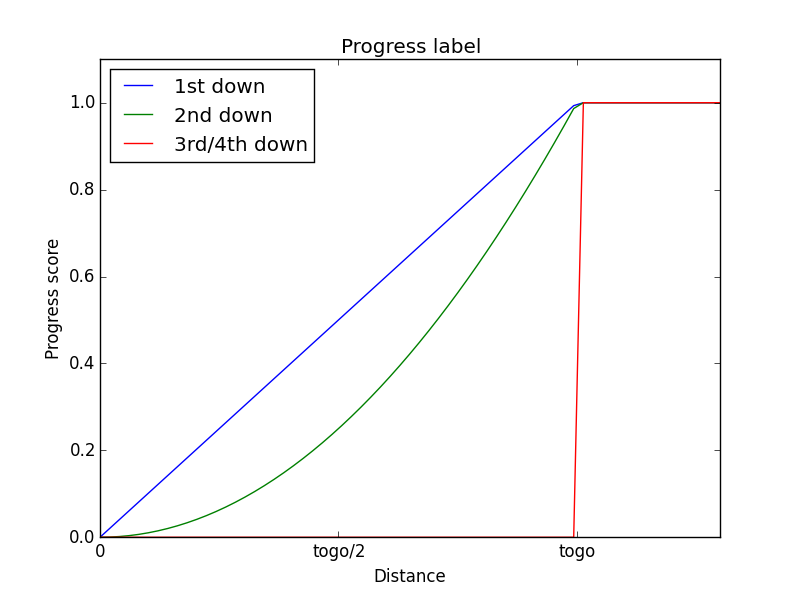}
  \caption{Progress measure}
  \label{fig:progress}
\end{figure}
The idea behind this is the following: If a new first down is not reached on third or fourth down it is (in most cases) a failure on the side of the offense and will result directly in a turnover on downs (in case of fourth down) or a punt (in case of third down). We neglect the case when teams are in field goal range and can still score because the ultimate goal is always to have a touchdown. These cases are all combined and labeled as 0 which represents failure. Any real number larger than zero represents success to some extent. The larger the number the better. A first down is still represented as 1. Since 0 is a failure and 1 is a first down, it intuitively makes sense to label progress towards a first down with a number between 0 and 1. Progress even after the first down does not get a larger value. If the first down was not reached, the value is dependent on the ratio of the remaining yards and the yards that were gained. Depending on the down this ratio can be squared. The idea behind this is as follows: Imagine the following two situations: (1) 5 yards were gained on first down. (2) 5 yards were gained on second down (after no gain on first down). Even (2) is solid and leaves only 5 yards for third down. The ratio values are therefore $\frac{5}{10}$ and $\frac{5}{10}$ for first and second down, respectively. However, the second value is as large as the first even though the remaining attempts are less. In order to fix this issue, we argue that whichever gain is achieved on second down should be penalized in some way since only one additional attempt is available assuming fourth down is used for a punt or field goal. After first down, two more attempts are left which puts the offense in a more comfortable situation. The actual \textit{progress} scores using a quadratic function for second down which are therefore $\frac{5}{10}=0.5$ and $(\frac{5}{10})^2=0.25$. Figure~\ref{fig:progress} illustrates the idea. While first down progress scores are awarded linearly, second down progress scores are calculated by a quadratic function. The closer one gets to a first down, the closer the progress values of the two get.\\
Our first idea of a progress measure increased linearly even after the first down for additional yards, but with the larger numbers even the average error increased dramatically. For example, a prediction of 1.0 on first down - or in other words a 10 yard gain - would result in an error of 8 if the play was actually a 90 yard play (9.0 real score) whereas an incomplete pass would only result in an error of 1 (0.0 real score). This is even when the play was actually a 10 yard pass and the receiver managed to break free. Intuitively, the positive prediction should get a relatively low error which is not the case. We argue that it's more about getting a first down or not instead of whether we make some yards more or less after the first down since it's the first down that keeps the offense on the field. Due to this reasoning we decided for a constant value of 1 for all play that reach a new first down.

\section{Methods}
\label{sec:methods}
This section shortly explains all methods we use for classification and regression.

\subsection{Classification Trees}
Classification Trees~[10] are binary trees where each internal node defines a rule for one of the input features which divides the data into two subsets aiming for the best split. In order to avoid over-fitting it is useful to define a maximum depth for the tree. We used balanced class weighting to handle the imbalanced data. That is, the weighting factors for the classes are automatically adjusted to their inverse frequencies in the input data.

\begin{figure}[!tbp]
  \centering
  \begin{minipage}[b]{0.59\textwidth}
    \includegraphics[width=\textwidth]{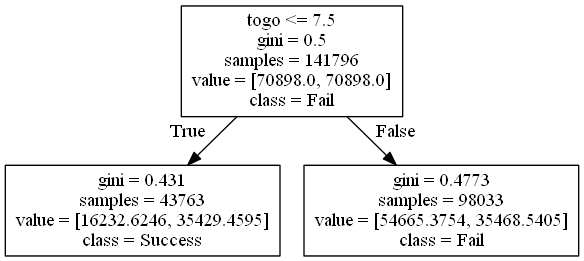}
      \caption{Decision tree with only one level.}
      \label{fig:dectree}
  \end{minipage}
  \hfill
  \begin{minipage}[b]{0.37\textwidth}
    \includegraphics[width=\textwidth]{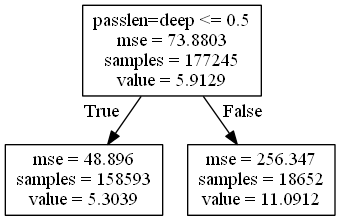}
  \caption{Regression tree with only one level.}
  \label{fig:regtree}
  \end{minipage}
\end{figure}

Figure~\ref{fig:dectree} shows a simple classification tree trained on our data set. It contains only a single decision rule which divides the data set into two subsets. The only decision rule here is whether the next first down is more than 7.5 yards away. That is, all plays where the team has 7.5 yards or less to go for their next first down are classified as "success" and all others as "failure". This simple rule already gives a surprisingly good accuracy of 69.2\% with 47.9\% precision and 50.0 \% recall. 

Table~\ref{tab:tree_results} in the Appendix shows the classification performance of classification trees with different depth limitations. The classification tree with just one rule achieves actually the best accuracy but its precision and recall are quite bad. The tree with a maximum depth of six achieves the highest F1 score.

\subsection{Regression Trees}

Regression trees are pretty similar to classification trees as they also use a set of rules to split the data into subsets. However, each final remaining set gets labeled with a real value.

Figure~\ref{fig:regtree} shows a simple regression tree trained on our data set for predicting the gained distance on a play. It contains only a single decision rule which divides the data set into two subsets. The only decision rule here is whether a deep pass has been played. That is, the first subset contains all plays with either a short pass or a run play and the second subset contains all plays with a deep pass. The first subset gets labeled with 5.3 yards and the second subset with 11.1 yards. With this decision rule we achieve a MAE of 5.7 yards and RMSE of 8.4 yards.

Table~\ref{tab:regtree_results} in the Appendix shows the performance of the regression tree with different depth limitations. It can be seen that the performance stays quite constant. However, using no depth limitation causes the algorithm to overfit the training data which results in poor performance.

\subsection{Nearest Centroid}
For Nearest Centroid classification we first calculate the centroids of the success and failure labels of the feature vectors in our training set. Then to predict the classification of a given feature vector we simply assign it to the class with the nearest centroid. This method provided approximately 50\% accuracy, precision, and recall after we undersampled the training set to have equal amounts of success and failure labels, so overall it is no better than a random assignment of the labels. This suggests that the classifications are not divided into two distinct clusters some distance apart, but are positioned in a way that makes distance from the centroids a poor indicator of class.

\subsection{Linear Discriminant Analysis}
For our Linear Discriminant Analysis classification we used Scikit Learn's LDAClassifer. We then compared the results between using Singular Value Decomposition, Least Squares, and Eigenvalue Decomposition to solve the LDA problem before classification. Both the Least Squares and Eigenvalue Decomposition solutions also used shrinkage to try and further improve the accuracy on our high dimensional feature vectors. All three solutions gave similar results with around 66\% classification accuracy and F1 scores.

\subsection{Support Vector Machines}
For support vector machines we first tried a simple linear decision boundary on an undersampled training set, and after performing a search over varying penalty values using 5-fold validation to compare performance. Our best result was 57.72\% accuracy at $C=2^{-5}$. We then tried using a Gaussian radial basis function as our kernel and performed a grid search over $C \in \{2^k | k \in [-5,17]\}$ and $\gamma \in \{2^k | k \in [-17, 4]\}$. We achieved our best accuracy of 66.65\% at $C = 2^11$ and $\gamma = 2^{-17}$. This accuracy is similar to what we achieved using the LDA Classifiers, but not an improvement.

\subsection{Support Vector Regression}
For regression we considered two different metrics: \textit{yards} to go, and \textit{progress}, which are discussed in the Section~\ref{sec:targets}. Once again we grid searched over values of $C \in \{2^k | k \in [-5,17]\}$ and $\gamma \in \{2^k | k \in [-17, 4]\}$ to try and find good hyperparameters for support vector regression.  Our best results were found at $C=2^7$ and $\gamma = 2^{-17}$ with a mean error of 5.207 yards and root mean squared error of 8.977 yards.  We were left dissatisfied by the performance in this regard as most frequently a first down is less than 10 yards away, so an error of 5 yards could be very significant.  For this reason we continued using the progress metric, and obtained a mean error of 0.1351 and root mean squared error of 0.2332.  As a rough estimate you can consider 0.13 progress to be approximately 13\% of the distance to a first down. So with the progress metric we obtained a much more useful prediction in terms of mean error, although the root mean squared error of 0.2332 shows that there is still quite a bit of variance in our results.

\subsection{Artificial Neural Networks}
Our Artificial Neural Network implementation uses the PyBrain~[3] library. PyBrain allows the user to specify a number of parameters including the number of hidden layers, the number of units per hidden layer, the type of units in the hidden layers and the maximum number of epochs for training the network.\\
It is worth mentioning that the training of the Neural Networks took considerably longer than model fitting for any other method which is why less configurations were explored. Moreover, this could be a disadvantage for a scenario when the network has to be retrained in real-time between plays or drives. For our experiments it meant that re-training the networks after taking into account the imbalanced data set for success classification was not possible. Without that, the neural networks classified almost everything (or nothing) as a failure. \\
We only present values that were reasonable. Especially the networks with linear units in the hidden layers performed badly and were not used for further experiments. The results are shown in Tables~\ref{tab:neuralnet_progress_results}, \ref{tab:neuralnet_yards_results} and~\ref{tab:neuralnet_success_results} in the Appendix. The only measure where the neural networks performed equally well compared to other regression methods was yards with a MAE of 5.524 and a RMSE of 8.751.\\

\section{Evaluation and Comparison}
\label{sec:eval}
%TODO: Review the changes made based upon the new data

The complete result tables are in the Appendix. Tables~\ref{tab:classifier_results}, \ref{tab:regression_results_yards} and \ref{tab:regression_results_progress} contain a compressed version of the results for the \textit{success}, \textit{yards} and \textit{progress} measures, respectively. We report only the data of the configuration with the best results for each of the approaches.\\

\begin{table}[h]
\centering
\begin{tabularx}{.8\textwidth}{ | X | c | c | c  | c | }
  \hline
  \textbf{Method} & \textbf{Accuracy} &\textbf{Precision}&  \textbf{Recall}&  \textbf{F1}\\ \hline
  Linear SVM & 57.72\% & 58.74\% & 51.59\% & 54.93\% \\ \hline
  RBF SVM & 66.65\% & \textbf{67.62}\% & 63.75\% & 65.63\% \\ \hline
  SVD LDA & 66.91\% & 67.20\% & 65.05\% & \textbf{66.11}\%  \\ \hline
  Least Squares LDA & 66.70\% & 67.24\% & \textbf{64.97}\% & 66.08\%  \\ \hline
  Eigenvalue LDA & 66.69\% & 67.40\% & 64.48\% & 65.91\%  \\ \hline
  %SGDC & 55.03\% & 57.52\% & 38.05\% & 45.80\%  \\ \hline  %Effectively the same method as Linear SVM since we used Hinge loss
  Nearest Centroid & 50.74\% & 50.68\% & 50.98\% & 50.83\%  \\ \hline
  %Decision Trees &  58.63\% & 58.56\% & 58.70\% & 58.86\% \\ \hline %These were Brendan's results for DT
  Decision Trees &  \textbf{67.14\%} & 46.08\% & 66.90\% & 54.57\% \\ \hline %I believe these were Valentin's results for DT
\end{tabularx}
\caption{Success classification performance of different methods}
\label{tab:classifier_results}
\end{table}
When comparing the results for \textit{success} accuracy is not the main criterion. Consider the real-world scenario where a team uses an application based on ones of these methods to select the next play. We have to make sure that plays that would result in a failure are classified as such, even if it is at the cost of classifying successful plays as failures. This is represented by precision or, in other words, the portion of true positives among all positives. High recall and accuracy, on the other hand, are nice to have but not as important. Based on this observation Support Vector Machines (SVM) using a Radial Basis Function (RBF) kernel is the best method with a precision of 67.62\%. In other words, in two thirds of the cases where this method predicts a play to result in a first down or touchdown the play actually is successful. Other methods perform similarly well and have a slightly higher recall and accuracy. Still, for applications this is most likely not enough.\\

\begin{table}[h]
\centering
\begin{tabularx}{.5\textwidth}{ | X | c | c | }
  \hline
  \textbf{Method} & \textbf{MAE} & \textbf{RMSE}\\ \hline
  RBF SVR & \textbf{5.207} & 8.977 \\ \hline
  Linear Regression & 5.569 & 8.309 \\ \hline
  Regression trees & 5.491 & \textbf{8.299}  \\ \hline
  Neural Network & 5.524 & 8.751 \\ \hline
\end{tabularx}
\caption{Yards regression performance of different methods}
\label{tab:regression_results_yards}
\end{table}
As an alternative to \textit{success} we also considered \textit{yards}. As expected, \textit{yards} is not a useful measure of success since it is prone to have large errors even for comparably good predictions, e.g. a 80-yard gain on a play with a predicted 30-yard gain results in a huge error. This is confirmed by the results in Table~\ref{tab:regression_results_yards}. The method with the lowest mean absolute error of 5.207 (MAE) is the RBF SVR. Considering that the distance to a first down is initially 10 yards, an error of more than 5 yards is quite substantial. The same applies to the root mean squared error (RMSE) of more than 8 for all methods. We conclude that \textit{yards} can not be used for a good application due to the high errors.\\

\begin{table}[h]
\centering
\begin{tabularx}{.5\textwidth}{ | X | c | c | }
  \hline
  \textbf{Method} & \textbf{MAE} & \textbf{RMSE}\\ \hline
  RBF SVR & \textbf{0.1351} & 0.2332 \\ \hline
  Linear Regression & 0.1412 & 0.2283 \\ \hline
  Regression trees & 0.1424 & \textbf{0.2131} \\ \hline
  Neural Networks & 0.1575 & 0.2449 \\\hline
\end{tabularx}
\caption{Progress regression performance of different methods}
\label{tab:regression_results_progress}
\end{table}
Finally, let us evaluate the results of the regression models using the \textit{progress} measure that we defined in Section~\ref{sec:targets}. Apart from the regression trees all of the other three methods performed fairly well. The best results were achieved by the RBF SVR with a MAE of 0.1351 and a RMSE of 0.2332. In order to understand the intuition let us consider the following example. For a typical 1st and 10 situation where no first down is achieved this would mean that our prediction is on average only 1.3 yards away from the real gain. For real predictions this is a very promising result. It is even more encouraging when we consider that third and fourth down situations still have potentially high errors since it is about predicting a binary variable. For these experiments, we did not round the third and fourth down predictions to 0 or 1 which might reduce the error even further.

\section{Applications}
\label{sec:applications}
Given that our methods evaluate the success of a play in a certain game situation one can go ahead and use all different plays a team has learnt. By that, the coach can determine which play is suited best in this particular game situation. Unfortunately, it is impossible to evaluate the effectiveness of this method based on our data set. We only have the result of the one play that was chosen, not the results of all plays that are in the team's playbook. The only way to measure whether it is indeed useful is to have actual teams use the method for a while and not use it for another time period and compare the success. Even then, one could argue that a number of factors other than the playcalling play a role including injuries, form, temperature, weather, etc.\\

%\section{Lessons learned}
%\label{sec:lessonslearned}
%This project has deepened our knowledge in Machine Learning in several ways. From learning how to deal with an imbalanced data set and how to successfully handle a mixture of categorical and continuous features in our data to applying statistical methods like PCA and ANOVA to better understand the data set so that we may devise better approaches to our problem, we have acquired skills that will be useful even when working on different tasks and data sets.\\
%Interestingly, we found that the definition of the targets is a major part to our success. As described in Section~\ref{sec:eval}, we were intrigued to see how the \textit{progress} measure can be predicted more accurately than others. \\
%Furthermore, we used a variety of classification and regression methods. From this experience we have not only learned how to apply them but also how different parameter settings affect the performance. 

%Better understanding of Classification methods
%Dealing with imbalanced data set
%Feature selection
%Labels for regression methods
%Categorical features

\section{Conclusion and Future Work}
\label{sec:conclusion}
In this work we aimed at providing predictions for the outcome of a specific play in a particular game situation in the NFL. This is a novel approach that could be used to identify the best play in real games by finding the available play with the maximum success before the play is actually executed.\\
In order to define a proper measure of success of a play we have come up with a new measure called \textit{progress} that is both intuitive and more accurately predictable than other measures at the same time. The accuracy we achieved by using a number of different regression models is promising for our intended application. Especially with more comprehensive data (formations, routes, players, etc.), more samples, advanced regression methods to get higher accuracy, and possibly by considering series of plays instead of single plays, the actual use in real-world playcalling is viable.\\
Furthermore, it could be useful to train classification or regression methods on plays of a specific team. The ANOVA statistic has shown that some teams are probably more predictable than others. This approach seems also more relevant for a real-world application since a coach is only interested in finding the best play for his own team.
The same approach could be used for a certain opponent. For example, when a team has a winning streak and seems to be unbeatable everyone is interested in finding out how to win against this team. \\

All of our source code is on GitHub~[1]. We explicitly encourage others to try using, modifying and extending it. Feedback and ideas for improvement are most welcome.\\
Finally, we would like to give credit to three open-source projects without which this work would not have been possible: scikit-learn~[4], PyBrain~[3] and nflgame~[2].

% use Chicago format
\section*{References}
\small{
[1] Lutz, Roman, Valentin Kassarnig and Brendan Teich. NFL Play Prediction repository at https://www.github.com/romanlutz/NFLPlayPrediction

[2] Gallant, Andrew. An API to retrieve and read NFL Game Center JSON data. https://github.com/BurntSushi/nflgame

[3] Schaul, Tom, Justin Bayer, Daan Wierstra, Yi Sun, Martin Felder, Frank Sehnke, Thomas R\"uckstie\ss, and J\"urgen Schmidhuber. "PyBrain." The Journal of Machine Learning Research 11 (2010): 743-746.

[4] Pedregosa, Fabian, Gaël Varoquaux, Alexandre Gramfort, Vincent Michel, Bertrand Thirion, Olivier Grisel, Mathieu Blondel et al. "Scikit-learn: Machine learning in Python." The Journal of Machine Learning Research 12 (2011): 2825-2830.

[5] McGarrity, Joseph P., and Brian Linnen. "Pass or run: an empirical test of the matching pennies game using data from the National Football League." Southern Economic Journal 76, no. 3 (2010): 791-810.

[6] Stilling, Stephanie T., and Thomas S. Critchfield. "The Matching Relation and situation-specific bias modulation in professional football play selection." Journal of the Experimental Analysis of Behavior 93, no. 3 (2010): 435-454.

[7] Kovash, Kenneth, and Steven D. Levitt. Professionals do not play minimax: evidence from Major League Baseball and the National Football League. No. w15347. National Bureau of Economic Research, 2009.

[8] Mitchell, Matthew. Dynamic Matching Pennies with Asymmetries: An Application to NFL Play Calling. Working Paper, 2010.

[9] Reed, Derek D., Thomas S. Critchfield and Brian K. Martens. "The generalized matching law in elite sport competition: Football play calling as operant choice." Journal of Applied Behavior Analysis 39, no. 3 (2006): 281.

[10] Breiman, Leo, Jerome Friedman, Charles J. Stone, and Richard A. Olshen. Classification and regression trees. CRC press, 1984.

[11] Molineaux, Matthew, David W. Aha, and Gita Sukthankar. "Beating the Defense: Using Plan Recognition to Inform Learning Agents." In FLAIRS Conference. 2009.

[12] Kahn, Joshua. "Neural network prediction of NFL football games." World Wide Web electronic publication (2003): 9-15.

[13] Purucker, Michael C. "Neural network quarterbacking." Potentials, IEEE 15, no. 3 (1996): 9-15.

[14] Sierra, Adrian, James Fosco, Carlos Fierro, and Vigilante TinMan Sleeping Tiger. "Football Futures." URL http://cs229.stanford. edu/proj2011/SierraFoscoFierro-FootballFutures.pdf.

[15] Min, Byungho, Jinhyuck Kim, Chongyoun Choe, Hyeonsang Eom, and RI Bob McKay. "A compound framework for sports results prediction: A football case study." Knowledge-Based Systems 21, no. 7 (2008): 551-562.

[16] Harville, David. "Predictions for National Football League games via linear-model methodology." Journal of the American Statistical Association 75, no. 371 (1980): 516-524.

[17] Pardee, Michael. "An artificial neural network approach to college football prediction and ranking." University of Wisconsin–Electrical and Computer Engineering Department (1999).

[18] Stefani, Raymond T. "Improved least squares football, basketball, and soccer predictions." IEEE transactions on systems, man, and cybernetics 10, no. 2 (1980): 116-123.

[19] Fokoue, Ernest, and Dan Foehrenbach. "A Statistical Data Mining Approach to Determining the Factors that Distinguish Championship Caliber Teams in the National Football League." (2013).

[20] Yu, Kai, and Tresp, Volker. "Heterogenous Data Fusion via a Probabilistic Latent-Variable Model." In Proceedings of 17th International Conference on Architecture of Computing Systems - Organic and Pervasive Computing (ARCS 2004): 20-30

[21] Lutz, Roman. "Fantasy Football Prediction", (2015) arXiv:1505.06918 [cs.LG]
}

\newpage

\section*{Appendix}

\begin{table}[H]
\begin{tabularx}{\textwidth}{ | X | l | l |}
  \hline
  \textbf{Description}& \textbf{Features} & \textbf{Labels} \\ \hline
  
 (9:56) M.Ryan pass short left to M.Jenkins to CAR 17 for 7 yards (T.Davis). &

  \Centerstack[l]{passlen: 'short' \\ time: 1496.0 \\ down: 2 \\ togo: 10 \\ half: 2 \\  position: 24 \\ team: 'ATL'\\ pass: 1 \\  side: 'left' \\  opponent: 'CAR'} 

  & 

 \Centerstack[l]{Success: 0 \\ Yards: 7.0 \\ Progress: 0.49} 

 \\ \hline
 
 (10:32) K.Moreno right tackle to OAK 27 for 1 yard (T.Kelly). &
  \Centerstack[l]{time: 1532.0\\ down: 2\\ togo: 1\\ half: 1\\ position: 28\\ team: 'DEN'\\ pass: 0\\ side': 'right'\\ opponent: 'OAK'} & 
 \Centerstack[l]{Success: 1 \\ Yards: 1.0\\ Progress: 1.0} \\ \hline
 
  (:27) (No Huddle, Shotgun) K.Warner pass deep left to L.Fitzgerald for 26 yards, TOUCHDOWN. &
  \Centerstack[l]{passlen: 'deep'\\ shotgun: 1\\ time: 27.0\\ down: 1\\ togo: 10\\ half: 1\\ position: 26\\ team: 'ARI'\\ pass: 1\\ side: 'left'\\ opponent: 'HOU' }
 & 
 \Centerstack[l]{Success: 1 \\ Yards: 26.0\\ Progress: 1.0} \\ \hline
\end{tabularx}
\caption{Feature Examples}
\label{tab:feature_examples}
\end{table}

\begin{table}

\begin{minipage}[b]{0.45\textwidth}
\centering
\begin{tabularx}{\textwidth}{ | X | r | }
  \hline
  \textbf{Feature} & \textbf{F-value} \\ \hline
   togo &	    		  16146.92 \\ \hline 
down &			      6690.42 \\ \hline 
pass &			      2927.44 \\ \hline 
passlen=short &		1256.44 \\ \hline 
passlen=deep &		867.02 \\ \hline 
shotgun &			    642.94 \\ \hline 
position &			  295.36 \\ \hline 
qbrun &			      87.01 \\ \hline 
time &			      56.66 \\ \hline 
team=NO &			    54.67\\ \hline 
team=NE &			    48.01 \\ \hline 
side=left &			  32.54 \\ \hline 
team=CLE &			  23.23\\ \hline 
team=DAL &			  22.78 \\ \hline 
opponent=NYJ &		22.67 \\ \hline 
team=STL &			  21.45 \\ \hline 
opponent=SF &			20.35 \\ \hline 
team=GB &			    19.44 \\ \hline 
team=SD &			    18.26 \\ \hline 
team=BUF &			  15.45 \\ \hline 
team=TEN &			  14.05 \\ \hline 
team=NYJ &			  13.90 \\ \hline 
team=TB &			    13.64 \\ \hline 
side=middle &			13.63 \\ \hline 
team=PIT &			  13.57 \\ \hline 
opponent=BAL &		13.15 \\ \hline 
team=OAK &			  13.12 \\ \hline 
team=JAC &			  13.11 \\ \hline 
team=ARI &			  12.61 \\ \hline 
team=PHI &			  10.11 \\ \hline 
team=DEN &			  8.76 \\ \hline 
opponent=TB &			8.72 \\ \hline 
opponent=NE &			8.56 \\ \hline 
team=KC &			    7.81 \\ \hline 
opponent=SEA &		7.45 \\ \hline 
opponent=PIT &		6.60 \\ \hline 
opponent=DAL &		6.11 \\ \hline 
opponent=ARI &		5.65 \\ \hline 
team=HOU &			  4.80 \\ \hline 
\end{tabularx}
\end{minipage} \qquad
\begin{minipage}[b]{0.45\textwidth}
\centering
\begin{tabularx}{\textwidth}{ | X | r | }
  \hline
  \textbf{Feature} & \textbf{F-value} \\ \hline
team=SF &			    4.43 \\ \hline 
opponent=IND &		3.87 \\ \hline 
opponent=ATL &		3.79 \\ \hline 
opponent=CIN &		3.67 \\ \hline 
half &			      3.50 \\ \hline 
side=right &			3.48 \\ \hline 
opponent=WAS &		3.23\\ \hline 
team=BAL &			  2.64 \\ \hline 
team=CIN &			  2.63 \\ \hline 
opponent=TEN &		2.56 \\ \hline 
opponent=NYG &		2.29 \\ \hline 
opponent=MIN &		2.24 \\ \hline 
team=ATL &			  2.21\\ \hline 
opponent=DEN &		1.97 \\ \hline 
opponent=STL &		1.92 \\ \hline 
opponent=CHI &		1.87\\ \hline 
opponent=JAC &		1.83 \\ \hline 
team=SEA &			  1.80 \\ \hline 
opponent=SD &			1.67 \\ \hline 
opponent=NO &			1.63 \\ \hline 
opponent=HOU &		1.57 \\ \hline 
team=IND &			  1.17 \\ \hline 
team=CHI &			  0.87 \\ \hline 
opponent=GB &			0.63 \\ \hline 
team=WAS &			  0.57 \\ \hline 
team=NYG &			  0.49 \\ \hline 
team=CAR &			  0.44 \\ \hline 
opponent=KC &			0.44 \\ \hline 
opponent=CAR &		0.38 \\ \hline 
opponent=MIA &		0.36 \\ \hline 
opponent=BUF &		0.10 \\ \hline 
opponent=CLE &		0.09 \\ \hline 
opponent=PHI &		0.09 \\ \hline 
team=DET &			  0.06 \\ \hline 
team=MIN &			  0.06 \\ \hline 
opponent=DET &		0.02 \\ \hline 
opponent=OAK &		0.001 \\ \hline 
team=MIA &			  0.00005 \\ \hline
\end{tabularx}
\end{minipage} 
\caption{Results of ANOVA F-test}
\label{tab:feature_scores}
\end{table}

\begin{table}
\centering
\begin{tabularx}{0.8\textwidth}{ | >{\centering\arraybackslash}X | c | c | c | c | }
  \hline
  \textbf{Maximum tree-depth} & \textbf{Accuracy} & \textbf{Precision} & \textbf{Recall} & \textbf{$F_1$}\\ \hline
  1 & \textbf{69.196\%} & \textbf{47.891\%} & 49.965\% & 48.906\%\\ \hline
  2 & \textbf{69.196\%} & \textbf{47.891\%} & 49.965\% & 48.906\%\\ \hline
  4 & 68.940\% & 47.847\% & 58.558\% & 52.663\%\\ \hline
  6 & 67.135\% & 46.079\% & \textbf{66.904\%} & \textbf{54.573\%}\\ \hline
  8 & 67.589\% & 46.521\% & 65.851\% & 54.524\%\\ \hline
  10 & 67.164\% & 46.054\% & 65.868\% & 54.207\%\\ \hline
  12 & 69.66\% & 45.798\% & 65.166\% & 53.792\%\\ \hline
  15 & 65.957\% & 44.703\% & 64.887\% & 52.936\%\\ \hline
  No limit & 64.321\% &  40.143\% & 42.607\% & 41.338\%\\ \hline
\end{tabularx}
\caption{Success classification performance of different Classification Tree configurations}
\label{tab:tree_results}
\end{table}

\begin{table}
\centering
\begin{tabularx}{0.8\textwidth}{ | >{\centering\arraybackslash}X | c | c | c | c |}
  \hline
  
   & \multicolumn{2}{c|}{\textbf{Yards}} & \multicolumn{2}{c|}{\textbf{Progress}}\\ \hline
  \textbf{Maximum tree-depth} & \textbf{MAE} & \textbf{RSME} & \textbf{MAE} & \textbf{RSME}\\ \hline
  
  1 & 5.7434 yds &8.4103 yds & 0.1667 & 0.2264 \\ \hline
  2 & 5.6762 yds & 8.3468 yds & 0.1508 & 0.2174 \\ \hline
  4 & 5.5134 yds & 8.2730 yds & 0.1451 & 0.2141 \\ \hline
  6 & 5.4943 yds & \textbf{8.2682 yds} & 0.1429 & \textbf{0.2131}   \\ \hline
  8 & \textbf{5.4906 yds} & 8.2991 yds & \textbf{0.1424} & \textbf{0.2131}  \\ \hline
  10 & 5.5137 yds & 8.3870 yds & \textbf{0.1424} & 0.2138  \\ \hline
  12 & 5.5520 yds & 8.5435 yds& 0.1427 & 0.2153  \\ \hline
  15 & 5.6350 yds & 8.8265 yds & 0.1436 & 0.2195 \\ \hline
  No limit & 7.2699 yds &  11.7828 yds &  0.1727 &  0.3026\\ \hline
\end{tabularx}
\caption{Yards and Progress performance of different Regression Tree configurations}
\label{tab:regtree_results}
\end{table}

\begin{table}
\centering
\begin{tabularx}{.6\textwidth}{ | >{\centering\arraybackslash}X | c | c |}
  \hline
  \textbf{Type} & \textbf{MAE} & \textbf{RSME}\\ \hline
  Tanh & \textbf{0.156087} & 0.247892 \\
  Sigmoid & 0.157487 & \textbf{0.244940} \\\hline
\end{tabularx}
\caption{Progress performance of different Neural Network configurations. In both cases, the number of hidden layers was 5 with 100 units each. The networks were trained over 100 epochs.}
\label{tab:neuralnet_progress_results}
\end{table}

\begin{table}
\centering
\begin{tabularx}{\textwidth}{| c | c | >{\centering\arraybackslash}X | c | c |}
  \hline
\textbf{hidden layers} & \textbf{hidden units} & \textbf{hidden class} & \textbf{RMSE} & \textbf{MAE}\\\hline
1 & 10 & Sigmoid & \textbf{8.595409} & 5.780438 \\ 
1 & 10 & Tanh & 8.595958 & 5.768907 \\ 
1 & 50 & Sigmoid & 8.599206 & 5.761803 \\ 
1 & 50 & Tanh & 8.621712 & 5.709222 \\ 
1 & 100 & Sigmoid & 8.611599 & 5.663897 \\ 
1 & 100 & Tanh & 8.650681 & 5.660227 \\ 
10 & 10 & Sigmoid & 8.595773 & 5.774482 \\ 
10 & 10 & Tanh & 8.599356 & 5.759645 \\ 
10 & 50 & Sigmoid & 8.595777 & 5.769012 \\ 
10 & 50 & Tanh & 8.618312 & 5.882702 \\ 
10 & 100 & Sigmoid & 8.599489 & 5.778361 \\ 
10 & 100 & Tanh & 8.751191 & \textbf{5.523861} \\ \hline
\end{tabularx}
\caption{Yards performance of different Neural Network configurations. All networks were trained for 100 epochs.}
\label{tab:neuralnet_yards_results}
\end{table}

\begin{table}
\centering
\begin{tabularx}{\textwidth}{ | >{\centering\arraybackslash}X | c | c | c | c | c | c | }
  \hline
  \textbf{hidden layers} & \textbf{hidden units} & \textbf{hidden class} & \textbf{Accuracy} & \textbf{Precision} & \textbf{Recall} & \textbf{$F_1$}\\ \hline
1 & 10 & Sigmoid & 0.704923 & 0.250000 & 0.000038 & 0.000076 \\ 
1 & 10 & Tanh & 0.704911 & 0.312500 & 0.000096 & 0.000191 \\ 
1 & 10 & Linear & 0.295055 & 0.295055 & \textbf{1.000000} & \textbf{0.455664} \\ 
1 & 50 & Sigmoid & 0.704911 & 0.486957 & 0.002142 & 0.004264 \\ 
1 & 50 & Tanh & 0.703112 & 0.363789 & 0.008299 & 0.016227 \\ 
1 & 50 & Linear & 0.295055 & 0.295055 & \textbf{1.000000} & \textbf{0.455664} \\ 
1 & 100 & Sigmoid & 0.704889 & 0.477477 & 0.002027 & 0.004037 \\ 
1 & 100 & Tanh & 0.702113 & 0.345062 & 0.010689 & 0.020736 \\ 
1 & 100 & Linear & 0.295055 & 0.295055 & \textbf{1.000000} & \textbf{0.455664} \\ 
5 & 100 & Sigmoid & \textbf{0.706598} & 0.593610 & 0.017764 & 0.034496 \\ 
5 & 100 & Tanh & 0.704945 & 0.500000 & 0.000057 & 0.000115 \\
10 & 10 & Sigmoid & 0.704945 & 0.000000 & 0.000000 & 0.000000 \\ 
10 & 10 & Tanh & 0.704939 & 0.000000 & 0.000000 & 0.000000 \\ 
10 & 10 & Linear & 0.295055 & 0.295055 & \textbf{1.000000} & \textbf{0.455664} \\ 
10 & 50 & Sigmoid & 0.705430 & \textbf{0.643333} & 0.003690 & 0.007339 \\ 
10 & 50 & Tanh & 0.704945 & 0.000000 & 0.000000 & 0.000000 \\ 
10 & 50 & Linear & 0.295055 & 0.295055 & \textbf{1.000000} & \textbf{0.455664} \\ 
10 & 100 & Sigmoid & 0.705199 & 0.574751 & 0.003308 & 0.006578 \\ 
10 & 100 & Tanh & 0.704945 & 0.000000 & 0.000000 & 0.000000 \\ 
10 & 100 & Linear & 0.295055 & 0.295055 & \textbf{1.000000} & \textbf{0.455664} \\ \hline
\end{tabularx}
\caption{Success classification performance of different Neural Network configurations. All networks were trained for 100 epochs.}
\label{tab:neuralnet_success_results}
\end{table}

\begin{table}
\centering
\resizebox{\textwidth}{!}{%
\begin{tabularx}{1.32\textwidth}{ |l | c | c | c | c | c | c | c | c | c | c | c | }
  \hline
 & \multicolumn{11}{c|}{\textbf{C}} \\ \hline 
$\gamma$ & \textbf{$2^{-5}$} & \textbf{$2^{-3}$} & \textbf{$2^{-1}$} & \textbf{$2^{1}$} & \textbf{$2^{3}$} & \textbf{$2^{5}$} & \textbf{$2^{7}$} & \textbf{$2^{9}$} & \textbf{$2^{11}$} & \textbf{$2^{13}$} & \textbf{$2^{15}$}\\ \hline 
$2^{-17}$ & 0.50169 & 0.51240 & 0.53899 & 0.60339 & 0.63748 & 0.64492 & 0.65033 & 0.65619 & \textbf{0.66323} & 0.66069 & 0.65816\\ \hline 
$2^{-15}$ & 0.50699 & 0.51668 & 0.55550 & 0.63483 & 0.63799 & 0.64650 & 0.65168 & 0.65996 & 0.65675 & 0.65095 & 0.63917\\ \hline 
$2^{-13}$ & 0.51003 & 0.52440 & 0.58733 & 0.63449 & 0.64227 & 0.65106 & 0.65405 & 0.65472 & 0.63940 & 0.61759 & 0.60018\\ \hline 
$2^{-11}$ & 0.52090 & 0.53978 & 0.60238 & 0.63438 & 0.64345 & 0.64864 & 0.64244 & 0.61906 & 0.58936 & 0.57528 & 0.56840\\ \hline 
$2^{-9}$ & 0.51983 & 0.55037 & 0.59235 & 0.61432 & 0.61697 & 0.61241 & 0.58818 & 0.57443 & 0.57387 & 0.57415 & 0.57392\\ \hline 
$2^{-7}$ & 0.50152 & 0.55956 & 0.57945 & 0.58835 & 0.58164 & 0.56936 & 0.56592 & 0.56558 & 0.56547 & 0.56547 & 0.56547\\ \hline 
$2^{-5}$ & 0.50096 & 0.50400 & 0.56294 & 0.55849 & 0.55544 & 0.55454 & 0.55426 & 0.55426 & 0.55426 & 0.55426 & 0.55426\\ \hline 
$2^{-3}$ & 0.49887 & 0.49921 & 0.51245 & 0.53882 & 0.53876 & 0.53876 & 0.53876 & 0.53876 & 0.53876 & 0.53876 & 0.53876\\ \hline 
$2^{-1}$ & 0.49887 & 0.49887 & 0.49893 & 0.49977 & 0.49977 & 0.49977 & 0.49977 & 0.49977 & 0.49977 & 0.49977 & 0.49977\\ \hline 
$2^{1}$ & 0.49887 & 0.49887 & 0.49887 & 0.50338 & 0.50338 & 0.50338 & 0.50338 & 0.50338 & 0.50338 & 0.50338 & 0.50338\\ \hline 
$2^{3}$ & 0.49887 & 0.49887 & 0.49887 & 0.50344 & 0.50344 & 0.50344 & 0.50344 & 0.50344 & 0.50344 & 0.50344 & 0.50344\\ \hline 
\end{tabularx}%
}
\caption{RBF SVM Accuracies for the 2014 season corresponding to different hyper-parameter settings.}
\label{tab:rbf_svm_params}
\end{table}

\end{document}